%% file: main.tex
\documentclass{sig-alternate-05-2015}

\usepackage{verbatim}
\usepackage{subfigure}
\usepackage{color}
\usepackage{float}
\usepackage{array}
\usepackage{booktabs}
\usepackage{graphicx}
\usepackage{amsmath}
\usepackage{amsfonts}
\usepackage{algorithm}
\usepackage{algpseudocode}
\usepackage{times}
\usepackage{balance}
\usepackage{tabu}
\usepackage{multirow}
\usepackage[flushleft]{threeparttable}

\newcommand{\hide}[1]{} 

\newcommand{\figref}[1]{Fig.~\ref{#1}} 

\newcommand*\samethanks[1][\value{footnote}]{\footnotemark[#1]}

\begin{document}

\setcopyright{acmcopyright}



\conferenceinfo{KDD'16}{August 13-17, 2016, San Francisco, CA, USA}


\title{Feature Engineering and Ensemble Modeling for Paper Acceptance Rank Prediction}
\numberofauthors{5}
\author{
Yujie Qian\thanks{indicates equal contribution} ,  Yinpeng Dong\samethanks\  , Ye Ma\samethanks\ , Hailong Jin, and Juanzi Li\\
	\affaddr{Department of Computer Science and Technology, Tsinghua University}\\
	\email{\{qyj13, dongyp13, y-ma13\}@mails.tsinghua.edu.cn,}\\
    \email{tsinghua\_phd@163.com, lijuanzi@tsinghua.edu.cn}
}

\maketitle


\input{abstract.tex}
\input{intro.tex}
\input{framework.tex}
\input{feature.tex}
\input{model.tex}

\input{exp.tex}
\input{conclusion.tex}

\section{ACKNOWLEDGMENTS}
The work is supported by 973 Program (No. 2014CB340504), NSFC-ANR (No. 61261130588), NSFC key project(No. 61533018), Tsinghua University Initiative Scientific Research Program (No. 20131089256).

\bibliographystyle{abbrv}
\bibliography{sigproc}

\end{document}

%% file: abstract.tex
\begin{abstract}
Measuring research impact and ranking academic achievement are important and challenging problems. Having an objective picture of research institution is particularly valuable for students, parents and funding agencies, and also attracts attention from government and industry. KDD Cup 2016 proposes the \textit{paper acceptance rank prediction} task, in which the participants are asked to rank the importance of institutions based on predicting how many of their papers will be accepted at the 8 top conferences in computer science. In our work, we adopt a three-step feature engineering method, including basic features definition, finding similar conferences to enhance the feature set, and dimension reduction using PCA. We propose three ranking models and the ensemble methods for combining such models. Our experiment verifies the effectiveness of our approach. In KDD Cup 2016, we achieved the overall rank of the 2nd place. 
\end{abstract}

%% file: intro.tex
\section{Introduction}
\label{sec:intro}

Mining academic data and academic social network attracts great research attention in a long time. Many issues in academic network have been investigated and several systems have been developed, such as DBLP, Google Scholar, Microsoft Academic Search, and Aminer~\cite{tang2008arnetminer}. One problem of importance and also difficulty in this field is to measure institution's academic achievement and research impact. Specifically, given a research area, such as Machine Learning, Data Mining, etc., how to rank the most influential institutions, like CMU, Stanford, and MIT?

KDD CUP 2016 focuses on this problem and proposes an innovative and interesting task: \textit{paper acceptance rank prediction}.  Given an upcoming top conference in 2016, the goal of this competition is to  rank the importance of institutions based on predicting how many of their papers will be accepted. In the competition, 8 computer science conferences are selected as target conferences, which are SIGIR, SIGMOD, SIGCOMM, KDD, ICML, FSE, MobiCom, and MM.

The competition's dataset includes the Microsoft Academic Graph (MAG) \cite{sinha2015overview}, and any other publicly available data on the Web. MAG is a large and heterogeneous academic graph provided by Microsoft, containing scientific publication records, citation relationships between publications, as well as authors, institutions, journal and conference venues, and fields of study. The latest version of MAG includes 19,843 institutions, 114,698,044 authors, and 126,909,021 publications.

The evaluation is performed after the conferences announce their decisions of paper acceptance. For every conference, the competition organizers collect the full list of accepted papers, calculate ground truth ranking, and evaluate the participants' submissions. Ground truth ranking is generated following a simple policy to determine the Institution Ranking Score:
\begin{enumerate}
\item Each accepted paper has an equal vote (i.e., they are equally important).
\item Each author has an equal contribution to a paper.
\item If an author has multiple affiliations, each affiliation also contributes equally.
\end{enumerate}
The evaluation are conducted in three phases, each chooses one conference to evaluate the predicted result. The competition uses NDCG@20 as the evaluation metric \cite{jarvelin2002cumulated}, which means only the top 20 institutions will be considered in the evaluation.

There have been some related works of this year's KDD Cup. A few studies try to solve the problem of  ranking different objects in academic network, e.g., authors, publications, conferences, and institutions. Christian Zimmermann summarized academic ranking problems and existing methods in \cite{zimmermann2013academic}. Various ranking criteria can be used in these rankings such as number of works, citation counts, h-index, impact factors, and aggregation of different methods. Besides, many learning-based approaches have been proposed to solve the general ranking problems, which are known as learning to rank \cite{liu2009learning}.

However, this competition is still novel and challenging. First, different from previous KDD Cup challenges, the ground truth (i.e., paper acceptance in 2016) is unknown beforehand, and with considerable uncertainty. Second, the participants do not necessarily use supervised learning algorithms since the problem is an open problem.

In this paper, we introduce the framework of our approach in the competition. We concretely describe our feature engineering methods, including basic feature definitions, finding similar conferences, and dimension reduction. We propose three ranking models, and use the ensemble of different models for final prediction. We also conduct several experiments to verify the effectiveness of our approaches. In KDD Cup 2016, we finally get an overall ranking of the 2nd place, while 10th in Phase 1 (SIGIR), 39th in Phase 2 (KDD), and 14th in Phase 3 (MM).

%% file: framework.tex
\section{Framework}
\label{sec:framework} 

\begin{figure}
\centering
\includegraphics[width=2.6in]{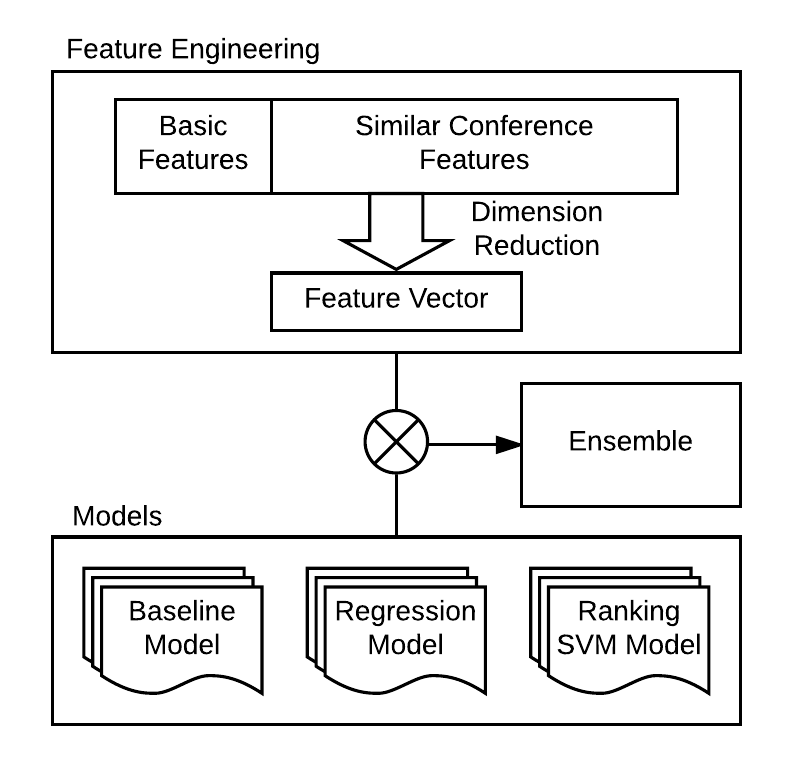}
\caption{\label{fig:framework}Framework of our solution.}
\end{figure}

The framework of our solution is illustrated in \figref{fig:framework}. It mainly consists of three components: feature extracting and preprocessing, model selection and training, model blending and ensemble. Details of these three parts will be elaborated in the ensuing sections. Below are some general discussion.

The first part of our framework is feature engineering, which is considered to be the most fundamental stage of data mining tasks. We first analyze the given dataset MAG, build up database for future use, and define certain basic and intuitive features. Then we are looking for methods to expand our collection of features, particularly by finding similar conferences to each targeted conference. Finally, we conduct dimension reduction through PCA in order to gain better performance.

In the second part, we primarily select three ranking models in our experiments, including a simple baseline model based on the ranking score in history, popular regression models such as linear regression and support vector regression, and some learning to rank models such as Ranking SVM. We train our models in the training set, and tune the hyper-parameters in the validation set.

In the final part, we apply blending methods to our models, which can effectively integrate different models and make them share complementary advantages.

%% file: feature.tex
\section{Feature Engineering}
\label{sec:feature} 

In this section, we introduce the feature set we derived in our work. Our feature engineering process mainly consists of three steps. At first, we define several basic features for each Institution-Conference-Year tuple. When ranking an institution in a conference, we use the features of the institution in the conference at recent 3 years. Then, we propose an approach to find similar conferences to each interested conference, and add these similar conferences in the feature set in order to incorporate more useful information. Finally, we conduct dimension reduction on the feature set to improve the performance of the ranking models. 

\subsection{Basic Features}
We define six basic features for each Institution-Conference-Year tuple. Specifically, for a tuple of Institution $A$, Conference $B$, Year(s) $C$, these features indicates Institution $A$'s performance at Conference $B$ in Year(s) $C$. The features are listed in Table \ref{feature}. All the features we used are statistical features, representing the institution's performance at the conference. We count the number of first and second authors because the first author is the main contributor of the paper, and the second author is usually the mentor, or another important author. We also calculate the Institution Ranking Score following the metric defined by KDD Cup organizers.

\begin{table}[t]
\caption{\label{feature}Features defined for each Institution-Conference-Year tuple $(A, B, C)$.}
\centering
\begin{tabular}{c|m{2.2in}}
\toprule
\textbf{Feature} & \textbf{Discription} \\\midrule
\#paper & Number of papers published \\\hline
\#author & Number of authors who published at least one paper \\\hline
\#author-paper & Number of (author, paper) pair \\\hline
\#1st author & Number of first authors \\\hline
\#2nd author & Number of second authors \\\hline
score & Institution Ranking Score, as described in Section \ref{sec:intro} \\
\bottomrule
\end{tabular}
\end{table}

When training ranking model for a conference, we generate features for each institution in last three years, each year separately and four years together. Finally, we have a basic feature vector of length 24 for each institution.

\subsection{Similar Conference Features}
In order to expand our collection of features and utilize plenty of other information, we find the most similar and representative conferences for each given conference and extract their features to supplement the existing ones. Similar conference features are helpful to predict paper acceptance in a targeted conference. Because of the uncertainty of paper submission and acceptance of an institution, only the targeted conference itself is not enough for prediction. It is common that an institution's performance in a particular conference varies from year to year. Features extracted from similar conferences can help to make more comprehensive and stable measure. 

The method we used to find similar conferences is based on collaborative filtering. Concretely, for each one of the eight given conferences, we follow the procedures below:

First of all, we traverse the database to generate the Author-Conference matrix $\mathbf{A}$, satisfying condition that $\mathbf{A}_{ij}=1$ if and only if author $i$ has published papers on conference $j$. During the computation, two extra points are considered. First, we only take recent published papers (no earlier than 2010) into account, which is essential for improving the timeliness of our results. Second, the row number (i.e., author number) of matrix $\mathbf{A}$ equals to the number of authors who has published papers on the given conference, instead of the number of all authors. This can effectively reduce the memory overhead and omit useless information.

Secondly, after getting the author-conference matrix $\mathbf{A}$, we examine two methods to compute similarity. The first one is to apply L1 normalization on matrix and compute cosine similarity of different column (i.e., conference) vectors. The second method is more intuitive. We just sum each column up to a row vector and find the maximum ones, which indicates they are more similar to the given conference on account of paper number published by authors who has published papers on the given conference.

The results of the similar conferences computed by our algorithm are listed in Table \ref{similar}. From the table, we can find out the results are basically coincide with our knowledge, such as KDD is most similar with ICDM, ICML is most similar with NIPS, etc.

It is worth mentioning that we determine the similar conferences directly from data instead of manually assigning, which is more convincing and can better capture the correlation between conferences. For example, we find that CVPR is similar to ICML, but actually CVPR is a conference on computer vision while ICML is a conference on machine learning. It is because computer vision researchers also publish many papers in machine learning conferences, such as their theoretical works such as new models and improved algorithms. This kind of correlation is informative for prediction and can be captured by our method.

In this competition, we choose top 3 similar conferences for each targeted conference to enrich the feature set. For similar conferences, we still use the basic features defined in the last subsection to represent the institutions' performance.
 
\begin{table}[t]
\centering
\caption{Top five similar conferences for the eight targeted conferences}
\label{similar}
\begin{tabular}{c|>{\centering}m{0.6in}|>{\centering}m{0.6in}|>{\centering}m{0.6in}|m{0.6in}<{\centering}}
\toprule
\textbf{No.} & \textbf{KDD}     & \textbf{ICML}    & \textbf{SIGIR} & \textbf{SIGMOD} \\ \midrule
1 & ICDM             & NIPS             & CIKM           & ICDE            \\ 
2 & CIKM             & AAAI             & WWW            & VLDB            \\ 
3 & WWW              & CVPR             & ECIR           & CIKM            \\ 
4 & AAAI             & KDD              & WSDM           & KDD             \\ 
5 & ICDE             & ICASSP           & KDD            & EDBT            \\ \bottomrule \multicolumn{4}{c}{}\\ 
\toprule
\textbf{No.} & \textbf{SIGCOMM} & \textbf{MobiCom} & \textbf{FSE}   & \textbf{MM}     \\ \midrule
1 & InfoCom          & InfoCom          & ICSE           & ICME            \\ 
2 & ICC              & ICC              & ASE            & ICIP            \\ 
3 & GlobeCom         & GlobeCom         & ISSTA          & CVPR            \\ 
4 & NSDI             & SIGCOMM          & ICSM           & ICASSP          \\ 
5 & IMC              & MobiSys          & MSR            & ICCV            \\ \bottomrule
\end{tabular}
\end{table}

\subsection{Dimension Reduction}
\label{subsec:dimension}
Data with high dimension will cause the curse-of-dimensionality problem and degrade the efficiency of algorithms. Dimension reduction can mitigate the curse-of-dimensionality and other undesired problems, as illustrated in \cite{jimenez1998supervised}. Many algorithms for dimensionality reduction have been proposed \cite{van2009dimensionality}. Among them, the linear algorithm Principal Component Analysis (PCA) \cite{pearson1901liii} is the most popular because of its effectiveness. 

PCA dimension reduction has the following steps:
\begin{enumerate}
\item Minus the empirical mean;
\item Compute the covariance matrix $S = \frac{1}{N}\sum_n\mathbf{x}_n\mathbf{x}_n^T$;
\item Eigenvalue decomposition. Let V denote the eigenvectors of the top d eigenvalues of S;
\item Reduce the dimension of data $Y = V^TX$.
\end{enumerate}

In the \textit{paper acceptance rank prediction} task, the number of affiliations is relatively small (741) and many of them are duplicates (many affiliations have the same feature vector, the elements of which all equal to zero), which makes it susceptible to over-fitting problem. To overcome this flaw, we use PCA algorithm to reduce the dimension of extracted features and obtain lower dimension features which are fed to later ranking or regression models. 

In order to determine the target feature dimension $K$, we use the most common criterion:
\begin{equation}
\frac{\sum_{i=1}^K\lambda_i}{\sum_{i=1}^N\lambda_i} > \tau
\end{equation}
where $N$ is the dimension of initial data space, $\{\lambda_1, \lambda_2, ..., \lambda_N\}$ are the eigenvalues of sample covariance matrix, and $\tau$ is a threshold which is usually 0.9 or 0.95. In this competition, we choose $\tau=0.95$ and find that $K = 5$ can satisfy all the requirements. So we perform PCA to obtain a 5-dimension feature vector as the input of ranking models.

%% file: model.tex
\section{Model}
\label{sec:model}

In this section, we introduce three ranking models and the ensemble methods we used in the competition. Each single model is suitable for this task, and we combine these three models for final prediction.

\subsection{Baseline Model}

The most straight-forward idea to predict an institution's paper acceptance in a conference this year is to measure its performance at this conference in the last few years. It is natural to assume that an institution which published a large number of papers at a conference last year or the year before last will still have many accepted papers at this year's same conference. According to this idea, we propose a baseline model.

The KDD Cup organizers defined Institution Ranking Score (as described in Section \ref{sec:intro}) for a conference, written as $Score_i^t$, represents institution $i$'s score in year $t$. Our baseline model uses the following metric to predict this year's score:
\begin{equation}
Pred_i^t = \frac{1}{\tau} \sum_{j=1}^{\tau} Score_i^{t-j}
\end{equation}
i.e., using the average Institution Ranking Score in the last $\tau$ years as the prediction score for this year. In the competition, we choose $\tau=5$. Small $\tau$ emphasizes the most recent years, but can be unstable if a institution have an occasional success or failure in a recent year. Large $\tau$ takes more years into consideration, but cannot distinguish the institution whose productivity is increasing or declining.

The baseline model follows a simple ranking criterion and does not need any training data. This model works well when the institutions have stable performance in the conference, since it uses average score and does not consider changes over years.

\subsection{Regression Model}
In supervised learning, the most popular method is regression. The goal of regression is to predict one or more continuous target variables $\mathbf{Y}$ given $D$ dimensional feature vector $(x_1,  x_2, ...,x_D)$ as input variables. The simplest regression model is linear regression which involves a linear combination of input variables, 
\begin{equation}
y(\mathbf{x}, \mathbf{w}) = w_0 + w_1x_1 + ... + w_Dx_D
\end{equation}
where $\mathbf{x} = (x_1,  x_2,.. ,x_D)$. The training goal is to learn the set of parameters $\mathbf{w} = (w_0, w_1,...,w_D)$. We can perform maximum-likelihood estimation for linear regression, which is equivalent to minimize the sum-of-squares error \cite{bishop2006pattern}, defined as
\begin{equation}
E_D(\mathbf{w})=\frac{1}{2}\sum_{n=1}^N(y_n-\mathbf{w^T\phi(x_n)})^2.
\end{equation}

Then we can find the optimal parameter $\mathbf{w^*}$ using stochastic gradient descent (SGD) method.

In this task, we treat the ranking score as continuous target variable and the extracted feature illustrated in Section \ref{sec:feature} as input variables. Regression model takes the feature vector as input, and directly output the predicted ranking score in 2016 for each institution. We use linear regression model and support vector regression with linear kernel, and average the output of two models to get the final result. The output of regression model is the ranking score for each institution, and then we can easily obtain the rank of each institution.

\subsection{Ranking SVM Model}
Learning to rank refers to the machine learning approaches of training models in a ranking task. Consider predicting paper acceptance as a ranking problem, we can apply some learning to rank models in this task. Existing learning to rank models can be categorized into three groups: pointwise, pairwise, and listwise approaches \cite{liu2009learning,hang2011short}. In this competition, we use a pairwise approach: Ranking SVM \cite{joachims2002optimizing}.

Pairwise approach transforms the learning-to-rank problem into a classification problem -- given a pair of items, learning a binary classifier to tell which one should be ranked higher. Then the goal is to minimize average number of inversions in ranking. In Ranking SVM, we train SVM as the classifier.

Now we formally describe Ranking SVM. Suppose the training data is given as $\{(x_i^{(1)}, x_i^{(2)}, y_i)\}$, $i= 1,\dots, m$, where each instance contains two feature vectors $x_i^{(1)}$ and $x_i^{(2)}$, and a label $y_i \in \{+1, -1\}$ indicates which feature factor should be ranked ahead. $m$ is the size of training data.The learning task is to solve a Quadratic Problem,
\begin{equation}
\begin{split}
\min_{w, \xi} & \quad \frac{1}{2} || w ||^2 + C\sum_{i=1}^m \xi_i \\
\text{s.t.} & \quad y_i \langle w, x_i^{(1)} - x_i^{(2)} \rangle \geq 1- \xi_i \\
& \quad \xi_i \geq 0 \\
& \quad i = 1,\dots, m,
\end{split}
\end{equation}
where $||\cdot ||$ denotes $L2$-norm, and $C>0$ is a coefficient. It is equivalent to a non-constrained optimization problem,
\begin{equation}
\min_w \sum_{i=1}^{m} \max ( 0, 1 - y_i \langle w, x_i^{(1)} - x_i^{(2)} \rangle ) + \lambda ||w||^2
\end{equation}
where $\lambda=\frac{1}{2C}$.

We train the ranking model using historical data. For example, we can pretend to predict the rank in 2015 in the training process since we have already know the answer, and use earlier years' data as input. Pairwise approach (e.g., Ranking SVM) is more appropriate in this specific task, since we have limited number of ranked lists, but enough items in each ranked list.

\subsection{Ensemble}
Model Ensemble can enhance the overall performance of individual models. In classification problems,  the error rate of classifiers can often be reduced by bagging \cite{breiman1996bagging} which is a common method of model ensemble. The final model is the combination of many classifiers by uniform voting. 

In ranking problems, we can also use the idea of bagging. We train a set of ranking models $\mathbf{M} = (M_1, M_2, ..., M_K)$ and model $M_i$ gives a prediction ranking score $\mathbf{s_i} = (s_i^1, s_i^2, ..., s_i^n)$, where  $s_i^j$ is the score of instance $j$, and $n$ is total number of ranking instances. Note that each model's output should be normalized into $[0, 1]$. The ensemble method we use is to average the output scores of all the models, while the final prediction is given by
\begin{equation}
\mathbf{s} = \frac{1}{K}\sum_{i=1}^K\mathbf{s_i}
\end{equation} 

Ensemble modeling can give stabler ranking score compared with single models. Single model each has its own advantages and disadvantages, and probably has unforeseen problems such as over-fitting. Ensemble modeling can blend different models and give more reliable output.

%% file: exp.tex
\section{Experiments}
\label{sec:exp} 

Various experiments were performed to evaluate the performance of the proposed methodologies and well-designed features. More experimental analyses on the effectiveness of some components in our framework are also given. We determine the parameters for our ranking models in the experiments and then predict the paper acceptance ranking in 2016.

\subsection{Experimental Setup}

To demonstrate the effectiveness of our proposed method, we divide the dataset into training set and validation set. Since MAG dataset only contains the full paper list of targeted conferences from 2011 to 2015, we train our model by predicting paper acceptance in 2014 (using 2011-2013 data to generate input features and using 2014 score as ground truth), and validate our model by predicting paper acceptance in 2015 (using 2012-2014 data to generate input features and using 2015 score as ground truth). 

We train the ranking model for each targeted conference separately, i.e., SIGMOD, KDD and ICML will have different ranking models. As mentioned in Section \ref{sec:feature}, we define 6 basic features for each Institution-Conference-Year tuple. For each institution, we consider its performance in 6 conference settings: targeted conference (full paper), targeted conference (all paper), 3 similar conferences (all paper), and all these 4 conferences, and further in 4 year settings: last 3 years separately, and together. So the initial feature vector length for each conference will be 144 ($6\times 6\times 4$). High feature dimension is usually harmful to the model performance. So we further perform PCA algorithm and reduce the feature dimension to 5 (details can be found in Subsection \ref{subsec:dimension}). 

We compare the models introduced in Section \ref{sec:model}, i.e., Baseline, Regression (using the implementation of \textit{scikit-learn} toolbox \cite{scikit-learn}), Ranking SVM (using the implementation of SVMRank \cite{joachims2006training}), and Ensemble Modeling. We utilize the training data to train the model, and test the performance on the validation data. The same as the competition, we use NDCG@20 as the metric to measure the ranking quality. The definition of NDCG@$n$ is following,
\begin{equation}
\begin{split}
\text{DCG} @n=\sum_{i=1}^{n} \frac{Score_i}{\log_2^{i+1}}\\
\text{NDCG} @n=\frac{\text{DCG} @n}{\text{IDCG}@n}
\end{split}
\end{equation}
where IDCG@n is the DCG@n of the ideal rank.

When predicting the institution ranking in 2016, we use both training and validation data mentioned above to train the ranking models. Then we generate input feature vectors using 2013-2015 data, and use the ensemble of three models' output as our final prediction. 

All the experiment codes are implemented in C++, Python, and Matlab, and the evaluations are performed on an x86-64 machine with 2.70GHz Intel Core i5 CPU and 8GB RAM. The operating system is OS X 10.11.5.

\subsection{Experiment Results}

\begin{table}[t]
\caption{\label{result}Validation Results of Different Models and Ensemble (NDCG@20)}
\centering
\setlength\tabcolsep{3pt}
\begin{tabular}{c|>{\centering}m{0.58in}|>{\centering}m{0.58in}|>{\centering}m{0.58in}|m{0.58in}<{\centering}}
\toprule
\textbf{Conference} & \textbf{Baseline} & \textbf{Regression} & \textbf{Ranking SVM} & \textbf{Ensemble} \\\midrule
SIGIR & 0.7664 & 0.8101 & 0.8178 & \textbf{0.8189} \\\hline
SIGMOD & 0.7979 & 0.8179 & 0.8052 & \textbf{0.8222} \\\hline
SIGCOMM & 0.7788 &\textbf{0.8059} & 0.7738 & 0.8028 \\\hline
KDD & 0.7828 & \textbf{0.8341} & 0.8132 & 0.8325 \\\hline
ICML & \textbf{0.8918} & 0.8794 & 0.8688 & 0.8740 \\\hline
FSE & 0.5815 & \textbf{0.6145} & 0.5427 & 0.6050 \\\hline
MobiCom & 0.6763 & 0.6644 & 0.6820 & \textbf{0.6955} \\\hline
MM & 0.5331 & \textbf{0.5345} & 0.5344 & 0.5265 \\ \midrule
Average & 0.7261 & 0.7451& 0.7297 & \textbf{0.7472} \\
\bottomrule
\end{tabular}
\end{table}

\begin{figure*}[t]
\centering
\includegraphics[height=1.6in]{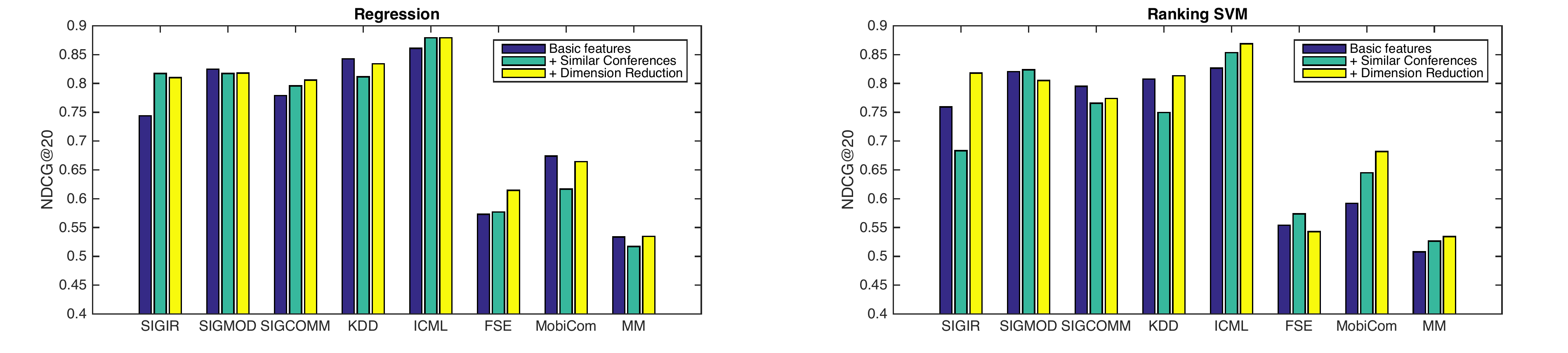}
\caption{\label{figdiscuss}Experiment results at different steps of feature engineering.}
\end{figure*}

We evaluate the performance using validation set. The results are shown in Table \ref{result}. From this table, we can see that regression model gets the highest NDCG@20 score in SIGCOMM, KDD, FSE and MM, while model ensemble achieves the highest score in SIGIR, SIGMOD and MobiCom. In average, model ensemble has  the best performance. Model ensemble does not  significantly outperform single model because these ranking models can achieve high performance themselves, and the number of models is also very limited. However, the ensemble of different models can reduce the uncertainty and get more reliable prediction compared with single model.

We note that the baseline method performs well enough compared with regression and ranking models. But the average of the last 5 years' score is very simple approach, and it cannot capture the trend of each institution's research ability. For example, if an institution gets a high score in the first years and decrease year by year, while another institution starts with relative low score, but keeps increasing recently. The baseline model may predict these two institutions the same, but we believe the latter one will do better because of the rising trend. Even though baseline model outperforms other methods in ICML prediction, we still choose the ensemble model to generate our final prediction.

\subsection{Discussions}

In the subsection, we investigate the effectiveness of  the three feature engineering steps in our framework. We report the results at different settings, starting from using basic features, and gradually improve the feature set. All the experiments are conducted on the same training and validation set as mentioned before. The performance of baseline model will not be included in this subsection because baseline model does not use any training data, so its prediction is independent of feature engineering methods.

\begin{table}[t]
\caption{\label{discusstabel}Performance comparison at different steps of feature engineering (average of 8 targeted conferences, NDCG@20). }
\centering
\setlength\tabcolsep{3.3pt}
\begin{tabular}{c|c|>{\centering}m{0.6in}|m{0.6in}<{\centering}}
\toprule
\textbf{No.} & \textbf{Method} & \textbf{Regression} & \textbf{Ranking SVM} \\ \midrule
1 & Basic features & 0.7289 &  0.7078  \\ \hline
2 & 1 + Similar conferences features & 0.7289 & 0.7026 \\ \hline
3 & 2 + Dimension reduction &  \textbf{0.7451} & \textbf{0.7297} \\ 
\bottomrule
\end{tabular}
\end{table}

We compare the results in three settings:

\begin{enumerate}
\item Only use basic features to train the model;
\item Add the similar conference features, but do not perform reduce dimension;
\item Add the similar conference features, and then use PCA to reduce dimension as mentioned in Section \ref{subsec:dimension}.
\end{enumerate}

Figure \ref{figdiscuss} shows the performance under each setting at 8 targeted conferences, and we list the average results in Table \ref{discusstabel}. 
We hope that adding similar conference features and performing dimension reduction can help with the prediction. However, the improvement does not always hold at every conference because of the uncertainty of paper acceptance. In average, we see that simply adding similar conference features have no improvement in regression model, and even a decrease of NDCG in Ranking SVM model. It is reasonable, because the dimension of feature vector expands a lot after adding similar conferences, and high dimension will cause overfitting. After performing dimension reduction, the NDCG significantly gains in both models. So we can conclude that similar conference features are beneficial for prediction as long as dimension reduction is performed.

The above experiments confirm the strengths of our proposed methods in the prediction task.

%% file: conclusion.tex
\section{Conclusion}
\label{sec:conclusion} 

In this paper, we introduce our solution at KDD Cup 2016 competition. We describe our effort in feature engineering, includes basic feature definition, finding similar conferences, and dimension reduction methods. We propose three ranking models and the ensemble method for prediction. Our empirical experiments illustrates the effectiveness of our approaches. In the competition, we achieved a final rank of the 2nd place.